%% file: main.tex
\documentclass[a4paper, 10pt, conference]{ieeeconf}      
\usepackage{FG2027}
\usepackage{tablefootnote}
\usepackage{multirow}
\usepackage{graphicx}
\usepackage{subcaption}
\usepackage{times} 
\usepackage{amsmath} 
\usepackage{amssymb}  

\FGfinalcopy 

\IEEEoverridecommandlockouts                

\def\FGPaperID{****} 

\title{\LARGE \bf
Pose-based Co-speech Gesture Detection
}

\author{\parbox{16cm}{\centering
    {\large Frances Yung$^1$ and Ninlawat Phuangchoke$^2$}\\
    {\normalsize $^1$ Language Science and Technology, Saarland University, Saarbr\"ucken, Germany \\
    $^2$ Information and Communication Technologies, Asian Institute of Technology, Klong Luang, Thailand}}
}

\begin{document}

\ifFGfinal
\thispagestyle{empty}
\pagestyle{empty}
\else
\author{Anonymous FG2027 submission\\ Paper ID \FGPaperID \\}
\pagestyle{plain}
\fi
\maketitle

\begin{abstract}

Co-speech gesture detection identifies the spans in which a speaker gestures (e.g., beat or semantic gestures) in video. As a first step in gesture annotation, it is crucial for studying human communicative behaviour. Despite the availability of automatic tools and models, co-speech gesture detection remains underexplored, and its current state of the art is unclear: existing methods have been evaluated using different datasets, experimental setups, and metrics, making their performance difficult to compare. To address this gap, we benchmark existing detectors using the same cross-validation folds across three datasets (MULTISIMO, GESRes, and CABB-S), with evaluation at the frame, gesture-span, and window levels. Our results show that previously reported performance does not consistently transfer to this common evaluation setting. Performance varies substantially across datasets, particularly with gesture frequency, indicating that the task remains far from solved. We further propose a CTR-GCN + BiLSTM model that outperforms the previous best method, MPSL, on CABB-~S. Our results provide a more consistent basis for comparing existing approaches and highlight key challenges for future research on co-speech gesture detection. 

\end{abstract}

\input{Sections/01-Introduction}

\input{Sections/02-related_work}

\input{Sections/03.A-data}
\input{Sections/03.B-task}

\input{Sections/03.C-proposal}
\input{Sections/04-experiment}
\input{Sections/05.results}
\input{Sections/06-analysis}
\input{Sections/07-conclusion}

\section*{ETHICAL IMPACT STATEMENT}
We collect no new data and use three existing datasets under their distributors' terms; consent and ethical review were handled by the original studies. We release only the data splits (fold time stamps) and the keypoints tracked from the original videos. Gesture detectors could be misused to monitor people without consent and, given their frequent errors, could bias research or assessments. To mitigate these risks, we report error rates per dataset and recommend human verification of detected gestures.

{\small
\bibliographystyle{ieee}
\bibliography{egbib}
}

\end{document}

%% file: Sections/01-Introduction.tex
\section{Introduction}

Co-speech gestures are an essential means of communication: performed alongside speech, hand and arm movements emphasise what is said or convey meaning that complements it \cite{kendon1988gestures,mcneill1992hand,goldin2013gesture}. How people produce and interpret gestures with speech is still largely an open question, and answering it at scale requires large annotated corpora. Annotating gestures in video, however, is labour-intensive, typically taking a human annotator several hours per hour of video.

The first step in gesture annotation is to identify gesture spans in continuous video \cite{Kita1998movement}. Several pose-based detectors have reported promising results, including a gesture recall of .87 for SPUDNIG \cite{ripperda2020speeding} and frame recalls of up to .98 for LightGBM \cite{ienaga2022semi} (Table~\ref{tab:comparison}). However, a more recent LightGBM-based tool reported a recall of .69 \cite{pouw2026envisionhgdetector}. Since these methods were evaluated on different data, often in-house, with different objectives and metrics, their results are not directly comparable, leaving the state of the art unclear and the practical reliability of existing detectors uncertain.

In this paper, we address this gap with the following contributions. (1)~We build a framework that evaluates existing gesture detection methods on the same data and with the same objectives: rule-based, pre-trained and trained methods are run on the same cross-validation folds of three datasets, MULTISIMO, GESRes and CABB-S, and scored per frame, per gesture span and per window. (2)~We show that the early reported success could not be replicated, because performance varies strongly across datasets, above all with how frequent gestures are. (3)~We show that models trained on the target data remain the most reliable, while cross-domain training and co-training with silver labels help only as far as the training data matches the test data. (4)~We propose a CTR-GCN + BiLSTM model that outperforms the previous best method, MPSL \cite{ghaleb2024co}, on CABB-S.

%% file: Sections/02-related_work.tex
\section{Related Work}

\subsection{Co-speech gestures}
Co-speech gestures are an integral part of human communication and convey information that complements and, in some cases, extends spoken language \cite{mcneill1992hand,kendon2004gesture}. For example, \textit{beat} gestures can emphasize or highlight parts of an utterance, while \textit{deictic} gestures direct the listener's attention toward specific locations or objects \cite{kendon1995gestures}. In contrast to conventionalized \textit{emblems}, such as the OK sign, which can carry relatively stable meanings independently of speech \cite{kendon1988gestures}, co-speech gestures are typically interpreted in conjunction with their accompanying speech and communicative context \cite{kendon1980gesture,kendon2004gesture}. For instance, the metaphoric gesture of presenting one palm and then the other can convey the contrast expressed by ``on the one hand, on the other hand'' only in combination with the corresponding speech. 

Compared with conventionalized gestures, co-speech gestures are more idiosyncratic and variable in form, reflecting differences across 
speakers and communicative contexts \cite{mcneill1985so,kendon2004gesture}.
The relationship between gesture and speech is not necessarily one-to-one: gestures may convey information that is not explicitly expressed in speech or may provide complementary information \cite{mcneill1994communicative,mcneill2005gesture}. For example, a shoulder shrug and open hands accompanied by an utterance of ``sure'' suggest the speaker is actually unsure.

Despite the importance of co-speech gestures, relatively few studies in computer vision have addressed their interpretation (see, e.g. \cite{hegde2026recognizing,zhang2026multimodal,nishida-etal-2025-multimodal}).
Understanding how humans produce and interpret gestures, on the other hand, remains an active area of research in the study of human communication and behavior (e.g. \cite{wagner2014gesture, holler2019multimodal}). Such studies commonly rely on manual annotation of gesture recordings \cite{Kita1998movement,bressem2024systems,Barros_2025}. A fundamental step in this process is identifying the temporal boundaries of gesture units in continuous motion streams. Each identified span can then be annotated with its gesture type (e.g., \textit{beat}, \textit{iconic}, or \textit{deictic}), kinematic properties (e.g. hand position, handshape, and trajectory), phases (e.g., preparation, stroke, hold, and retraction), and communicative function or meaning.  Accurate gesture-unit detection is therefore essential for subsequent annotation and analysis.

\subsection{Co-speech gesture detection}
Manual annotation of co-speech gestures is time-consuming, limiting the scalability of gesture analysis to larger datasets. 
Nonetheless, automatic gesture annotation remains largely unexplored\footnote{A related task is the identification of phonological components in sign language as part of sign recognition models \cite{kezar-etal-2023-improving,kezar2023exploring}. However, the applicability of these approaches to co-speech gestures has yet to be explored.} and only a limited number of studies have investigated the automatic detection of co-speech gesture units. 

\begin{table*}[htpb]
    \centering\small
    \setlength{\tabcolsep}{3pt}
    \renewcommand{\arraystretch}{1.1}
    \begin{tabular}{@{}p{0.11\textwidth}|p{0.14\textwidth}|p{0.145\textwidth}|p{0.10\textwidth}|p{0.09\textwidth}|p{0.08\textwidth}|p{0.115\textwidth}|p{0.12\textwidth}@{}}
         \hline
         method & task and default parameters  & data & train /dev& test& tracking & evaluation & gesture-R/P/F1 \\
         \hline
         SPUDNIG algorithm \cite{ripperda2020speeding}
         & per-frame prediction; removes spans $<4$ frames and merges spans $\le 4$ frames apart
         & in-house Dutch conversation videos
         & tuned on 207~held-out gesture spans
         & 20~videos $\times$~2~mins
         & OpenPose
         & recall, prec., Cohen's $\kappa$ per gesture \textbf{span},
         & 0.87 / 0.30 / - \\

         \hline

        \multirow[c]{2}{=}{SPUDNIG algorithm; reported in \cite{ivanova2026evaluation}}
       & \multirow[c]{2}{=}{(same as above)}
        & MULTISIMO\cite{koutsombogera2018modeling}: English collaborative group conversations
        & tracking confidence threshold tuned on 6 held-out videos
        & 38 selected videos $\times$~9~mins 
        & OpenPose
        & recall, prec., F1 per gesture \textbf{span}, Tversky index per gesture \textbf{frame}
        & 0.73/ 0.44/ 0.50\\
         \cline{3-6}
        \cline{8-8}

        && ECOLANG\cite{gu2025ecolang}: English interaction with object manipulation
        & tuned on 8 held-out videos
        & 88 selected videos $\times$~5 mins
        & OpenPose
        && 0.92/ 0.61/ 0.71\\
        \hline

        \multirow[c]{2}{=}{Velocity peaks algorithm \cite{ivanova2026evaluation}}
        & \multirow[c]{2}{=}{per-frame prediction;  removes segments $<7.5$ frames}
        & MULTISIMO\cite{koutsombogera2018modeling}: English collaborative group conversations
        & tuned on 6 held-out videos
        & 38 selected videos $\times$~9~mins 
        & DeepLab-Cut
        & recall, prec., F1 per gesture \textbf{span}, Tversky index per gesture \textbf{frame}
        & 0.68/ 0.43/ 0.48\\
         \cline{3-6}
        \cline{8-8}

        && ECOLANG\cite{gu2025ecolang}: English interaction with object manipulation
        & tuned on 8 held-out videos
        & 88 selected videos $\times$~5 mins
        & MediaPipe
        && 0.48/ 0.52/ 0.48 \\

        \hline

         \multirow[c]{2}{=}{LightGBM - 144 features \cite{ienaga2022semi}}
         &\multirow[c]{2}{=}{prediction per 0.5~s stride-1 sliding window; smoothing of segments $\le 0.2$~s}
         & in-house controlled monologues in Italian
         & 6~videos $\times$ 5~mins.
         & \multirow[c]{2}{=}{10-~fold CV. train/dev/test: 80/10/10}
         & \multirow{2}{=}{OpenPose}
         & \multirow{2}{=}{recall per gesture \textbf{frame}, accuracy and macro-F1}
         & 0.98 / 0.98 / - \\
         \cline{3-4}
         \cline{8-8}

         &  & in-house naturalistic monologues in Italian
         & 195~videos $\times$~27~secs.
         &&&& 0.87/ 0.89/ -\rule[-3.5ex]{0pt}{0pt}\\
        \hline

        ENVISION-LGBM \cite{pouw2026envisionhgdetector} 
        & prediction per 5-frame stride-1 sliding window; removes spans $<0.5$~s, merges spans $\le 0.3$~s apart
        & \multirow[c]{2}{=}{combination of 7 datasets: SAGA\cite{lucking2010bielefeld}, SAGA++\cite{Kucherenko2022multimodal}, MULTISIMO\cite{koutsombogera2018modeling}, ZHUBO \cite{bao2024editable}, TEDM3D \cite{rohrer2022temporal},
        ECOLANG\cite{gu2025ecolang}, GESRes \cite{hensel2025richly}}
        & \multirow[c]{2}{=}{7,774 gesture units, down-sampled non-gesture units}
        & \multirow[c]{2}{=}{train/dev/test: 60/10/30}
        & \multirow[c]{2}{=}{MediaPipe}
        & \multirow[c]{2}{=}{recall, precision, F1 per gesture/non-gesture \textbf{window}}
        & 0.69/ - /0.76 \\

        \cline{1-2}
        \cline{8-8}

        ENVISION-CNN \cite{pouw2026envisionhgdetector} 
        & prediction per 25-frame stride-1 sliding window; same smoothing parameters
         &&&&&& 0.56/ - / 0.52\rule[-5ex]{0pt}{0pt} \\
        \hline

        Multi-phase sequence labelling  (MPSL) \cite{ghaleb2024co}
         & per 18-frame sliding window prediction
         & \multirow[c]{2}{=}{CABB-S \cite{rasenberg2023mutual}: referential tasks in Dutch}
         & \multirow[c]{2}{=}{19~videos = total 16~hours, down-sampled non-gesture units}  
         &  \multirow[c]{2}{=}{5-fold~CV. train/test: 80/20}
         &  \multirow[c]{2}{=}{MMPose} 
         & recall, prec., F1, and Intersection over Union (IoU) per gesture \textbf{window}
         & 0.63/ 0.72/ 0.67\\
         \cline{1-2}
         \cline{7-8}

         Skeletal speech fusion \cite{ghaleb2024leveraging}
         & per 15-~frame sliding window prediction
         & & &&
         & F1 and Mean Averaged Precision (MAP) per gesture \textbf{window}
         & - / - / 0.70 \\
         \hline

    \end{tabular}
    \caption{Summary of major co-speech gesture detection methods proposed in the literature. The table reports the most commonly used evaluation metrics—\textbf{R}ecall, \textbf{P}recision, and \textbf{F1}-score—for the \textit{gesture} class, as reported by each study at its respective level of evaluation granularity.}
    \label{tab:comparison}
\end{table*}

At first glance, co-speech gesture detection may appear straightforward as it only involves a binary classification between gesturing and non-gesturing periods. Early approaches reported encouraging results. For example, SPUDNIG \cite{ripperda2020speeding} is a heuristic-based method that detects movements relative to estimated rest positions. A recall-based agreement against manual annotation of up to 0.92 was reported, although precision was low \cite{ripperda2020speeding,ivanova2026evaluation}. In addition, \cite{ienaga2022semi} trained a LightGBM classifier \cite{ke2017lightgbm} on body keypoint features, achieving a macro-F1 score of up to 0.98. In an active learning setting, the authors reported that an F1 score of at least 0.85 could be achieved by annotating 27\% of the data.

In contrast, recent models reported more moderate performance. \cite{pouw2026envisionhgdetector} introduced a gesture-detection tool, EnvisionHGdetector (ENVISION), that is trained on multiple gesture-annotated corpora using LightGBM and CNN models. Their reported F1 scores ranged from 0.52 to 0.76 on the combined evaluation set. \cite{ghaleb2024co} formulated gesture detection as a multi-phase sequence-labeling (MPSL) task, combining a Spatio-Temporal Graph Convolutional Network (ST-GCN) specialized for sign language recognition \cite{yan2018spatial,jiang2021skeleton} with a transformer. In subsequent work, they investigated strategies for integrating skeletal features with speech representations extracted from mel-spectrograms  \cite{ghaleb2024leveraging}. Both methods achieved a similar window-level F1 score of about 0.70.

Table \ref{tab:comparison} summarizes major studies on co-speech gesture detection in the literature. Overall, these studies establish the feasibility of learning-based gesture detection. However, their task objectives, datasets, and evaluation protocols largely differ.
A systematic benchmark is necessary to enable a fair comparison of existing approaches and to assess their respective strengths and limitations.

%% file: Sections/03.A-data.tex
\section{Methodology}
The objective of this study is to assess the current state of the art in automatic co-speech gesture detection. To this end, we systematically evaluate existing open-source methods under a common experimental setup, using multiple datasets and consistent evaluation protocols. Specifically, we compare four existing approaches: the rule-based SPUDNIG algorithm \cite{ripperda2020speeding}, the trainable LightGBM model proposed by \cite{ienaga2022semi}, the pretrained ENVISION framework \cite{pouw2026envisionhgdetector}, and the MPSL framework proposed by \cite{ghaleb2024co}. Furthermore, we introduce a new model based on the Channel-wise Topology Refinement Graph Convolutional Network (CTR-GCN) \cite{chen2021channel} to investigate whether techniques that have proved effective in skeleton-based action recognition can also improve co-speech gesture detection.

This section first introduces the datasets used in the experiment, followed by the definition of the detection task and the introduction of our proposed model.

\subsection{Data}
Among the limited datasets with manual gesture annotations, we selected three for the current study: MULTISIMO \cite{koutsombogera2018modeling}, GESRes \cite{hensel2025richly} and CABB-S~\cite{rasenberg2023mutual}.
The MULTISIMO corpus contains 23 videos (approximately 4 hours) of discussion on questions in a similar format as the game \textit{Family Feud}. We use videos of both the participants and the moderators, who are recorded individually in a seated position. The spans and types of gesture units were fully annotated by \textit{Rater~1} and partially annotated by \textit{Rater~2}. We only use the annotations by \textit{Rater~1}. 
GESRes contains 12 videos by politicians, clinicians, and educators -- three speakers each. Each video was annotated from the beginning but only partially, covering approximately 3.5 of the 7 hours and resulting in 2373 coded gesture spans.
The videos originate from existing recordings captured from a front camera angle.

The MULTISIMO and GESRes datasets depend little on the visual context. For evaluation on general co-speech gesture detection, we prioritize them over interactions that rely heavily on task-specific environments, such as ECOLANG \cite{gu2025ecolang}, which involves manual manipulation with physical objects. However, we also experiment with CABB-S, a smaller-scale variant of the CABB corpus \cite{eijk2022cabb}. It contains 19 dyads (approximately 5 hours) of task-oriented interactions in Dutch, in which one participant instructs the other to identify complex 3D objects. Participants are recorded using front-facing cameras while standing.  This is the only dataset that allows for the implementation of the multi-phase sequence labelling model \cite{ghaleb2024co} and our proposed variant, which require fine-grained annotation of gesture strokes.

The original videos of all three datasets are openly available (unlike e.g. SaGA \cite{lucking2010bielefeld}), allowing for different markerless tracking techniques and facilitating reproducibility. The recording conditions are rather stable, compared with the frequent camera changes or cuts in other datasets such as TEDM3D \cite{rohrer2022temporal} and the EVA corpus \cite{mlakar2023understanding}), which may interfere with pose tracking.

%% file: Sections/03.B-task.tex
\begin{figure*}[htbp]
    \centering
    \includegraphics[width=0.9\textwidth]{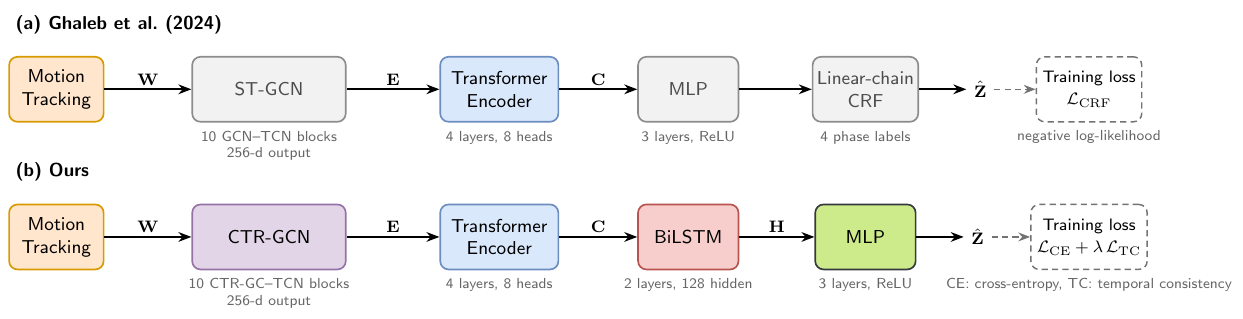}
    \caption{Comparison between the MPSL model of \cite{ghaleb2024co} and our proposal. ST-GCN is replaced by CTR-GC $+$ MS-TCN;
    CRF decoding by BiLSTM $+$ MLP; the Transformer encoder is unchanged.}
    \label{fig:model_diff}
\end{figure*}
\subsection{Task Definition}\label{sec:task}
To accommodate the different task formulations of the models included in the comparison, we define the co-speech gesture detection task at two levels of granularity, corresponding to two task objectives: per-frame prediction and per-window prediction.
\subsubsection{Per-frame prediction}
Given a sequence of two-dimensional body keypoints extracted from an RGB video using a pose-tracking method, the per-frame objective is to determine whether each frame belongs to a manually annotated gesture unit or to a non-gesture interval. Non-gesture intervals include both resting periods and non-gesture movements, such as nose scratching or headset adjustment. The present study considers visual input only and does not use speech signals as model input.
Let the input sequence be represented by:


\[
\mathbf{X} = (\mathbf{x}_1, \mathbf{x}_2, \ldots, \mathbf{x}_N)
\]

where $\mathbf{x}_n \in \mathbb{R}^{J\times 2}$ denotes the 2D coordinates of $J$ body keypoints at frame $n$, and $N$ is the number of frames in the video. The corresponding binary target sequence is:

\[
\mathbf{Y} = (y_1,y_2,\ldots,y_N),
\qquad
y_n \in \{0,1\}
\]
where $y_n=1$ indicates that frame $n$ belongs to an annotated gesture unit, and $y_n=0$ denotes a non-gesture frame. 

We evaluate SPUDNIG, ENVISION, and LightGBM on MULTISIMO and GESRes using the per-frame prediction
objective. All compared methods predict per sliding window, assign each
prediction to a frame, and smooth the resulting frame labels into gesture
spans with tool-specific minimum durations and gap thresholds
(Table~\ref{tab:comparison}). The spans are converted back to frame labels
on the normalised 29.97\,fps grid for evaluation.

\subsubsection{Per-window prediction}

The MPSL framework \cite{ghaleb2024co} adopts an alternative task design,
which optimizes for per-window prediction. The original data is preprocessed
into input sequences of $M=40$ sliding windows as:
\[
\mathbf{W} = (\mathbf{w}_1, \mathbf{w}_2,\ldots, \mathbf{w}_M)
\]\[
\mathbf{w}_m = (\mathbf{x}_{s_m}, \ldots, \mathbf{x}_{s_m+17}) \in \mathbb{R}^{18\times J\times 2}
\]

where $s_m$ is the first frame of window $m$ and consecutive windows are
shifted by two frames. Each input sequence corresponds to a sequence of
4-class labels, one for each window:
\[
\mathbf{Z} = (z_1, z_2, \ldots, z_M), \qquad z_m \in \mathcal{C} = \{P, S, R, N\}.
\]
These labels encode the degree and temporal location of a window's overlap
with a manually annotated gesture \textit{stroke}: partial overlap (less
than half of the window) at the end of the window ($P$), overlap with more
than half of the window ($S$), partial overlap at the beginning of the
window ($R$), or no overlap ($N$). They approximately correspond to the
\textbf{p}reparation, \textbf{s}troke, \textbf{r}etraction and
\textbf{n}eutral phases \cite{kendon1980gesture,Kita1998movement}, even though they are
defined operationally rather than by direct phase annotation.
The model is trained to predict the four-class sequence but evaluated
binarily, mapping the three non-neutral classes to \textit{gesture} and
$N$ to \textit{non-gesture}:
\[
y^{\mathrm{w}}_m = g(z_m) =
\begin{cases}
1, & z_m \in \{P,S,R\},\\
0, & z_m = N
\end{cases}
\]
and analogously $\hat{y}^{\mathrm{w}}_m = g(\hat{z}_m)$ for predicted labels.
Thus, the ground truth of a window is \textit{gesture} if any of its frames
overlaps an annotated gesture stroke, and the MPSL model produces one binary
prediction per 18-frame window. We evaluate this model on CABB-S, where
gesture strokes, rather than complete gesture units, are annotated. 

For comparison, the span outputs of all other models are first brought to the frame level.
The frame-level
predictions $\hat{y}_n$ are then aggregated over the same 18-frame
windows using an any-frame rule,
\[
\hat{y}^{\mathrm{w}}_m = \max_{s_m \le n \le s_m+17} \hat{y}_n ,
\]
i.e.\ a window is predicted as \textit{gesture} if any of its frames is
predicted as gesture.\footnote{Rule-based/pre-trained per-frame classification is based on overlap with a gesture
unit, while per-window classification is based on the gesture stroke, which
is only part of a unit. An aggregated per-window gesture prediction is
therefore a false positive if the window contains no stroke frame.}

%% file: Sections/03.C-proposal.tex
\subsection{Proposed method}
Our 
proposed method targets the per-window classification objective and builds on the MPSL model \cite{ghaleb2024co}, which embeds each skeletal
time window with an ST-GCN, contextualises the window sequence with a Transformer
encoder, and decodes labels with a linear-chain CRF. Our model keeps the general architecture, but replaces the ST-GCN and CRF components, as shown in 
Figure~\ref{fig:model_diff}.

Instead of ST-GCN, which convolves over a single fixed skeleton graph, we use
CTR-GCN \cite{chen2021channel}, which refines the joint topology separately per
channel. It contains a Multi-Scale Temporal Convolutional Network (MS-TCN) \cite{liu2020disentangling} that captures both
fine-grained and long-range patterns:

\[
    \mathbf{E} = \text{CTR-GCN}(\mathbf{W})
\]

In addition, the Transformer outputs 
$\mathbf{C}$
are not decoded with a CRF. Following \cite{cao2024advanced}, they are fed
to a BiLSTM \cite{graves2005framewise} that models dependencies between
gesture phases, and an MLP finally maps each position to class logits:
\[
\mathbf{H} = \mathrm{BiLSTM}\big(\mathbf{C}\big), \qquad
\hat{\mathbf{Z}} = \mathrm{MLP}\big(\mathbf{H}\big)
\]

In place of the
CRF objective ($\mathcal{L}_{\text{CRF}}$) \cite{fields2001probabilistic}, we minimize the cross-entropy loss ($\mathcal{L}_{\text{CE}}$) plus a temporal-consistency
loss ($\mathcal{L}_{\text{TC}}$) (weight $\lambda$ = 0.02). Adapted from prior temporal consistency frameworks \cite{lai2018learning, varghese2021unsupervised}, this loss penalizes prediction divergence between consecutive windows sharing the same ground-truth label while enforcing a sharp probability shift at class boundaries.

%% file: Sections/04-experiment.tex
\section{Experiment}
\subsection{Model setting}
The compared models differ in how, and on which data, they are trained.
SPUDNIG \cite{ripperda2020speeding} is rule-based and requires no training. ENVISION \cite{pouw2026envisionhgdetector} is used off the shelf with both of its released
backbones, a CNN and a LightGBM classifier; neither is retrained here. LightGBM \cite{ienaga2022semi} is
trained and evaluated separately on each dataset by cross-validation. We use the default parameter setting of each tool (see Table~\ref{tab:comparison}).

The MPSL model is re-trained using the implementation released with \cite{ghaleb2024co}. 

Similar to the MPSL model, our proposed model is also trained with the multi-phase objective, and then mapped to binary gesture labels per window. 

Training follows the setup of \cite{ghaleb2024co}: SGD
\cite{robbins1951stochastic} with Nesterov momentum of 0.9, a batch size
of 64, an initial learning rate of 0.1 and a weight decay of \(10^{-4}\),
for a fixed number of 80 epochs with a 20-epoch linear warm-up and a 10-fold learning-rate
decay at epoch 50. Due to architectural difference, our model is not initialized on the sign language GCN \cite{jiang2021skeleton}, but trained from scratch with a batch size of 16.

On top of training on the same corpus, we also tried a \emph{cross-domain} setting: we train LightGBM (MediaPipe features) on one dataset and test it on the other, i.e.\ train on all of GESRes and test on MULTISIMO, and vice versa. Furthermore, we use the cross-domain model to label the training-fold videos of the test dataset; these \emph{silver} labels, together with the out-of-domain data, are used to train a \emph{co-training} model, e.g.\ trained on GESRes + silver MULTISIMO and tested on MULTISIMO. These setups test how well a trained detector transfers to new data.

\subsection{Data preprocessing}

All videos are resampled to 29.97\,fps; lower-frame-rate recordings
(23.97\,fps in GESRes) are upsampled by frame interpolation. 
Each model uses the pose estimator it was originally developed with, which
has been found to outperform the alternatives \cite{ivanova2026evaluation}:
OpenPose \cite{cao2017realtime} (25 keypoints from BODY\_25) for SPUDNIG and
LightGBM; MediaPipe \cite{zhang2020mediapipe} (543 body, hand
and face landmarks) for ENVISION, and additionally for LightGBM as a
comparison; and MMPose \cite{mmpose2020} (133 keypoints) for the MPSL model and ours.
Each model requires a different subset of the tracked keypoints (see original papers for details).
We remove corrupt samples and tracking artifacts, and filter frames in which
static background equipment is detected as a person. Seven MULTISIMO
sessions are excluded because of tracking failures and
occlusions.\footnote{P019\_S08, M001\_S02, M001\_S11, M002\_S07, M002\_S08,
M003\_S22, M003\_S23.}

CABB-S provides per-window binary labels derived from stroke annotations. MULTISIMO and GESRes annotate each gesture-unit span with a
category labels. We treat spans labelled \textit{symbolic}, \textit{beat},
\textit{iconic}, \textit{deictic}, \textit{metaphoric} or \textit{emblem} as
gestures. All other spans, namely \textit{N/A} (MULTISIMO), \textit{adaptor}
(GESRes), \textit{non-gesture} (CABB-S) and unannotated spans, are treated as
non-gestures.

We construct speaker-disjoint cross-validation splits for MULTISIMO (4 folds) and
GESRes (3 folds). For CABB-S, the MPSL models use the preprocessed multi-phase data
and five folds of \cite{ghaleb2024co}, which are also split by speakers. LightGBM is trained on the original
per-frame annotations with the same folds. SPUDNIG and ENVISION are run on each complete dataset and each sample appears in exactly one test fold.
Table~\ref{tab:gold_distribution} shows that the gesture proportion varies
strongly across datasets (6\% in MULTISIMO vs.\ 35\% in GESRes) and across
folds (2--8\% in MULTISIMO, 17--54\% in GESRes). We will release our preprocessed and timestamped tracking data (the videos are available from their original sources) as a common benchmark.

\begin{table}[ht]
\centering
\begin{tabular}{@{}llrrr@{}}
\hline
fold & split & size & gesture& non-gesture \\
\hline
\multicolumn{2}{@{}l}{\textbf{MULTISIMO}}& frame count	\\	

0	&train	&	587,889	&    6\%&	94\%\\
0	& test	&	195,250	&	7\%&	93\%\\
1	&train	&	583,956	&	8\%&	92\%\\
1	&test	&	199,183	&	2\%&	98\%\\
2	&train	&	586,434	&	6\%&	94\%\\
2	&test	&	196,705	&	8\%&	92\%\\
3	&train	&	591,138	&	6\%&	94\%\\
3	&test	&	192,001	&	8\%&	92\%\\
\hline
\multicolumn{2}{r}{corpus size} & 783,139 & 6\% & 94\% \\
avg. & train/test & 75\%/25\% & & \\
\hline\hline
\multicolumn{2}{@{}l}{\textbf{GESRes}} & frame count	\\	
0 &	train   & 256,949   &	45\%&55\%\\
0	& test    &	148,361	&	17\%&	83\%\\
1	& train	&	272,082	&	34\%&	66\%\\
1	& test	&	133,228	&	37\%&	63\%\\
2	& train	&	281,598	&	26\%&	74\%\\
2	& test	&	123,721&	54\%&	46\%\\
\hline
\multicolumn{2}{r}{corpus size} & 405,310 & 35\% & 65\% \\
avg. & train/test & 67\%/33\% & & \\
\hline\hline
\multicolumn{2}{@{}l}{\textbf{CABB-S}} & window count	\\
0 & train & 662,440 & 10\% & 90\% \\
0 & test & 186,200 & 11\% & 89\% \\
1 & train & 662,800 & 11\% & 89\% \\
1 & test & 185,840 & 11\% & 89\% \\
2 & train & 677,280 & 11\% & 89\% \\
2 & test & 171,360 & 7\% & 93\% \\
3 & train & 711,680 & 10\% & 90\% \\
3 & test & 136,960 & 13\% & 87\% \\
4 & train & 680,360 & 10\% & 90\% \\
4 & test & 168,280 & 11\% & 89\% \\
\hline
\multicolumn{2}{r}{corpus size} & 848,640 & 11\% & 89\% \\
avg. & train/test & 80\%/20\% & & \\
\hline
\end{tabular}
\caption{Gesture and non-gesture proportions per fold}
\label{tab:gold_distribution}
\end{table}
\begin{figure}[htp]
    \centering
    \includegraphics[width=\linewidth]{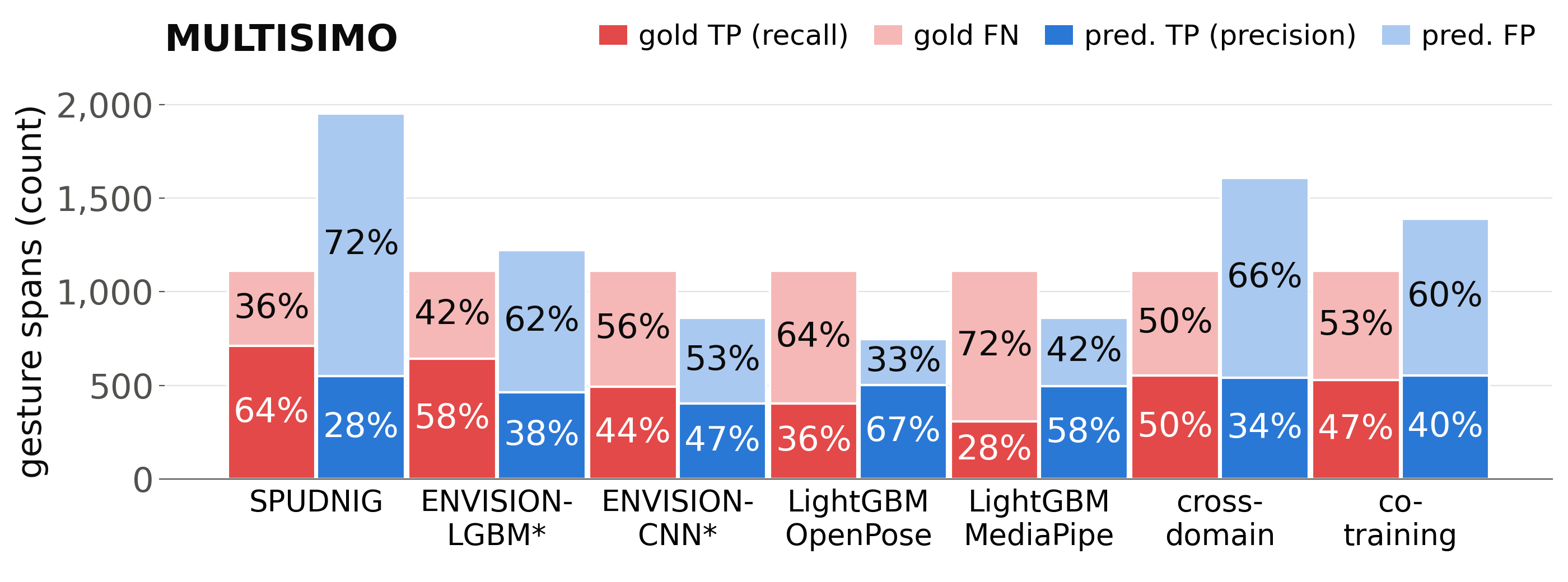}\\[0.3em]
    \includegraphics[width=\linewidth]{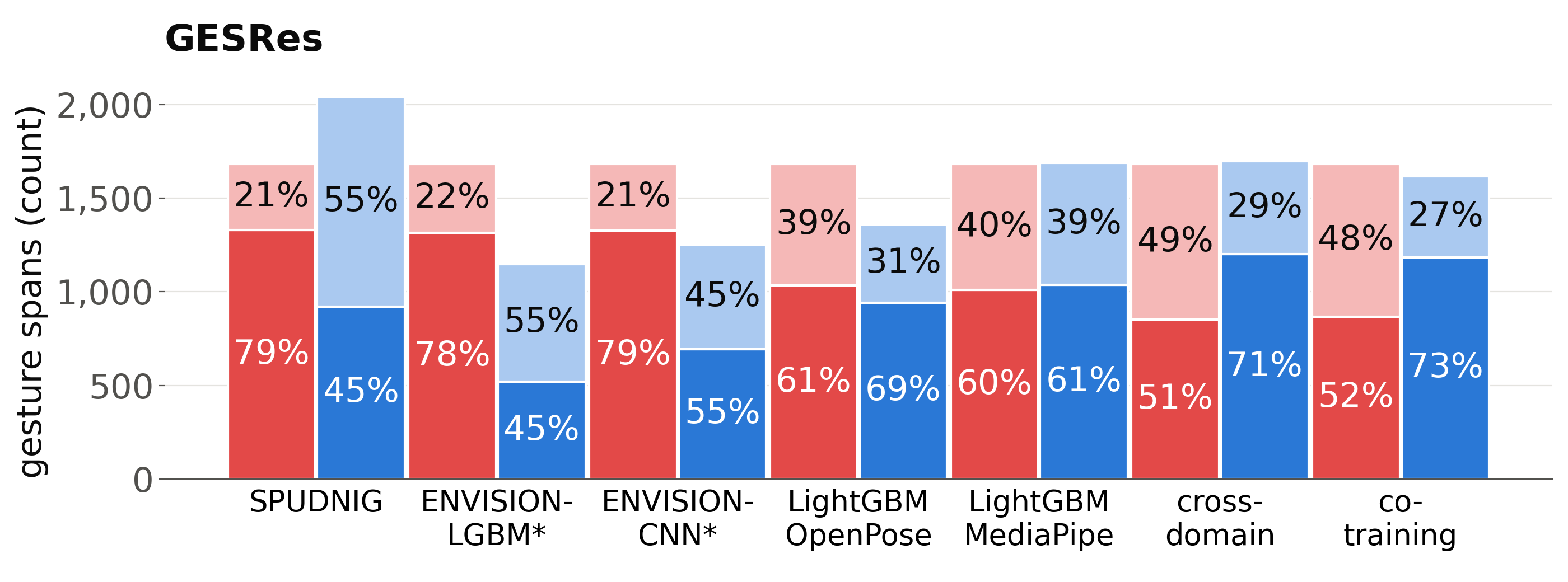}\\[0.3em]
    \includegraphics[width=\linewidth]{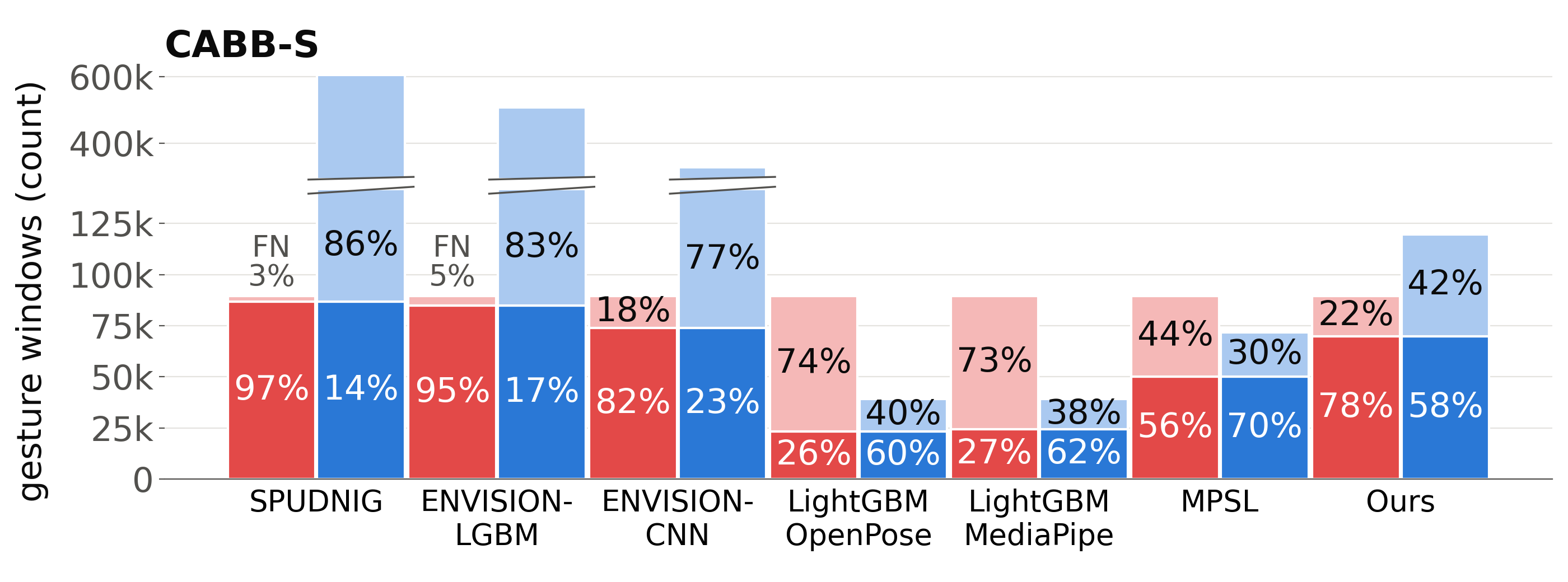}
    \caption{Gold vs.\ predicted gesture window (CABB-S) or span (MULTISIMO and GESRes) counts, summed over all test folds. The red bars show gold span counts, split into TP (share=~recall) and FN (missed). The blue bars show the predicted window/span counts, split into TP percentage (=~precision) and FP (unmatched). The matching on spans is not one-to-one, so the gold and predicted TP counts can differ. $^*$ Test data partly overlaps with training data.}
  \label{fig:predict_vs_gold_span_counts}
\end{figure}

%% file: Sections/05.results.tex
\begin{table*}[h]
    \centering \small
    \begin{tabular}{@{}l|l|l|ll|l|l@{}}
        \hline
        & & train & \multicolumn{2}{c|}{per frame} & per-span & gesture span \# \\
        method & tracking & training data & F1 & accuracy & F1 & (avg.\ length) \\
        \hline
        SPUDNIG & OpenPose & -- & .433 $\pm$ .13 & .904 $\pm$ .03 & .352 $\pm$ .13 & 1,953 (1.56) \\
        ENVISION-LGBM$^*$ & MediaPipe & external$^*$ & .461 $\pm$ .13 & .920 $\pm$ .03 & .426 $\pm$ .13 & 1,223 (2.04) \\
        ENVISION-CNN$^*$ & MediaPipe & external$^*$ & .419 $\pm$ .17 & .925 $\pm$ .01 & .422 $\pm$ .11 & 860 (2.24) \\
        \hline
        LightGBM & OpenPose & MULTISIMO & \textbf{.504 $\pm$ .06} & \textbf{.952 $\pm$ .02} & \textbf{.457 $\pm$ .04} & 747 (1.28) \\
        LightGBM & MediaPipe & MULTISIMO & .409 $\pm$ .13 & .946 $\pm$ .01 & .344 $\pm$ .10 & 862 (1.03) \\
        \hline
        cross-domain LightGBM & MediaPipe & GESRes & .397 $\pm$ .12 & .911 $\pm$ .02 & .372 $\pm$ .13 & 1,608 (1.50) \\
        co-training LightGBM & MediaPipe & + silver MULTISIMO & .439 $\pm$ .12 & .928 $\pm$ .02 & .398 $\pm$ .13 & 1,391 (1.37) \\
        \hline
    \end{tabular}
    \caption{Gesture detection on \textbf{MULTISIMO} (6.2\% gold gesture = 1,112 gold gesture spans, average 1.46\,s). Mean $\pm$ SD over the 4 CV folds. Per-frame F1 for the gesture class and accuracy over all frames. Predicted gesture span count is summed over the folds. Best results are bold, skipping the methods marked with $^*$, because the test data partly overlaps with the training data.}
    \label{tab:multisimo_metrics}
\end{table*}
\begin{table*}[htpb]
    \centering \small
    \begin{tabular}{@{}l|l|l|ll|l|l@{}}
        \hline
        & & training data & \multicolumn{2}{c|}{per frame} & per-span & gesture span \# \\
        method & tracking & data & F1 & accuracy & F1 & (avg.\ length) \\
        \hline
        SPUDNIG & OpenPose & -- & \textbf{.635 $\pm$ .11} & .708 $\pm$ .10 & .534 $\pm$ .20 & 2,041 (3.12) \\
        ENVISION-LGBM$^*$ & MediaPipe & external$^*$ & .673 $\pm$ .04 & .735 $\pm$ .10 & .561 $\pm$ .04 & 1,148 (5.42) \\
        ENVISION-CNN$^*$ & MediaPipe & external$^*$ & .656 $\pm$ .08 & .703 $\pm$ .14 & .645 $\pm$ .03 & 1,252 (5.48) \\
        \hline
        LightGBM & OpenPose & GESRes & .631 $\pm$ .09 & .762 $\pm$ .08 & \textbf{.625 $\pm$ .11} & 1,360 (3.01) \\
        LightGBM & MediaPipe & GESRes & .625 $\pm$ .05 & .752 $\pm$ .08 & .593 $\pm$ .06 & 1,689 (2.50) \\
        \hline
        cross-domain LightGBM & MediaPipe & MULTISIMO & .615 $\pm$ .02 & .761 $\pm$ .11 & .586 $\pm$ .03 & 1,697 (1.99) \\
        co-training LightGBM & MediaPipe & + silver GESRes & .623 $\pm$ .03 & \textbf{.765 $\pm$ .11} & .604 $\pm$ .02 & 1,618 (2.06) \\
        \hline
    \end{tabular}
    \caption{Gesture detection on \textbf{GESRes} (34.7\% gold gesture = 1,683 gold gesture spans, average 2.78\,s). Mean $\pm$ SD over the 3 CV folds. (See caption of Table~\ref{tab:multisimo_metrics})}
    \label{tab:gesres_metrics}
\end{table*}
\subsection{Evaluation}
On MULTISIMO and GESRes, we evaluate on the per-frame objective. The MPSL model and our proposed model are excluded, as they require training on stroke annotations. These two datasets are also part of ENVISION's training data, and since its train-test split is not publicly available, our test folds may overlap with its training data. Its scores on these two datasets may therefore be inflated and should be interpreted with caution.
LightGBM, in contrast, is always evaluated on folds held out from its training data. 

We additionally evaluate span-level performance by merging consecutive frame predictions into gesture and non-gesture spans. Following \cite{ripperda2020speeding}, a predicted span matches a ground-truth span if their temporal overlap is at least 60\%. Span-level evaluation captures the detection of individual gestures, regardless of duration. This evaluation better aligns with the objective of manual annotation, as short gesture spans are no less relevant than longer ones. We report precision, recall, and F1-score for the gesture class at the corresponding prediction level.
On CABB-S, all models are evaluated on the per-window objective: frame-level predictions are aggregated with the any-frame rule.
\section{Results}
\label{sec:results}
We first present the model performance on MULTISIMO and GESRes, evaluated on the same cross-validation folds\footnote{SPUDNIG and ENVISION's metrics are are computed per test fold and averaged.}.
Table~\ref{tab:multisimo_metrics} shows that MULTISIMO is the harder of the two datasets: no method exceeds a per-frame F1 of .504, which is achieved by LightGBM with OpenPose tracking, as in the original proposal \cite{ienaga2022semi}. On GESRes (Table~\ref{tab:gesres_metrics}), the per-frame F1 of models are all above .60. The two datasets differ most in their gesture ratio: 6.2\% in MULTISIMO against 34.7\% in GESRes.
The rule-based SPUDNIG is clearly below the trained LightGBM on MULTISIMO (.433 vs.\ .504 frame F1), but on GESRes the two are on par. ENVISION, which is trained on a variety of data partly including the test data, achieves superior scores only on GESRes, not on MULTISIMO. Overall, LightGBM has the most stable performance across both datasets.
%
%
%
%
\begin{table*}[h]
    \centering \small
    \begin{tabular}{@{}l|l|l|l|ll@{}}
        \hline
        & &  & gesture windows \# & \multicolumn{2}{c}{per window} \\
        method & tracking & training data & (sum of all folds) & F1 & accuracy \\
        \hline
        SPUDNIG & OpenPose & -- & 604,722 & .250 $\pm$ .04 & .387 $\pm$ .03 \\
        ENVISION-LGBM & MediaPipe & external & 508,387 & .286 $\pm$ .05 & .497 $\pm$ .07 \\
        ENVISION-CNN & MediaPipe & external & 327,616 & .357 $\pm$ .06 & .684 $\pm$ .08 \\
        \hline
        LightGBM & OpenPose & CABB-S (binary, per-frame)& 39,192 & .348 $\pm$ .09 & .902 $\pm$ .02 \\
        LightGBM & MediaPipe & CABB-S (binary, per-frame)& 39,179 & .372 $\pm$ .05 & .904 $\pm$ .02 \\
        \hline
        MPSL (SL-GCN + CRF) & MMPose & CABB-S (multi-class,   & 71,623 & .617 $\pm$ .03 & \textbf{.928 $\pm$ .01} \\
        Ours (CTR-GCN + BiLSTM) & MMPose & window sequences) & 119,709 & \textbf{.666 $\pm$ .03} & .918 $\pm$ .01 \\
        \hline
    \end{tabular}
   \caption{Gesture detection on \textbf{CABB-S}, evaluated per window
    (fixed 18 frames = 0.6\,s; 10.6\% gold gesture = 89,640 windows).
    Models grouped by training data. F1 is for the gesture class and
    accuracy over all windows; mean $\pm$ SD over the 5 cross-validation
    folds, using the model of the final epoch.
    Best results are bold.}
  \label{tab:cabbs_metrics}
\end{table*}
Figure~\ref{fig:predict_vs_gold_span_counts} explains the performance gap across datasets. The off-the-shelf tools over-detect gesture spans (taller blue bars) and favour recall over precision. This tendency is costly on MULTISIMO, where gestures are rare, but less penalised when gestures are frequent, as in GESRes. The LightGBM models, in contrast, generally under-predict (shorter blue bars) and favour precision over recall, which is an advantage when gestures are rare, as in MULTISIMO.

None of the existing methods reaches the scores originally reported for it (Table~\ref{tab:comparison}), even on the same dataset. For example, \cite{ivanova2026evaluation} reported a gesture span F1 of .50 for SPUDNIG on MULTISIMO, against .352 in our experiment. LightGBM was reported with frame recalls of .87 to .98 \cite{ienaga2022semi}, but on MULTISIMO its frame recall drops to .42. Tables~\ref{tab:multisimo_metrics} and \ref{tab:gesres_metrics} also show that performance varies considerably between test speakers of the same dataset (SD up to .17 in frame F1).

The \emph{cross-domain} models lag behind the in-domain versions, as expected, and their errors follow the gesture ratio of their training data. Trained on GESRes, where gestures are frequent, the model over-predicts on MULTISIMO; trained on MULTISIMO, where gestures are rare, it under-predicts on GESRes at the frame level, which is reflected in shorter detected gesture spans (1.99\,s vs.\ 2.78\,s for the gold spans). Co-training with silver labels of the target dataset reduces this mismatch on both datasets and achieves results comparable to LightGBM-MediaPipe, with the highest frame accuracy on GESRes (.765).

Next, we look at the results on CABB-S.
Table~\ref{tab:cabbs_metrics} shows the per-window results, where 10.6\% of the windows are gold gesture windows. Both SPUDNIG and ENVISION label most windows as gesture, resulting in a high recall (82--97\%) and a very low precision (14--23\%; Figure~\ref{fig:predict_vs_gold_span_counts}). ENVISION's reported window F1 of .76 (LGBM) and .52 (CNN) \cite{pouw2026envisionhgdetector} is thus not reached on CABB-S. LightGBM, which tends to under-detect gestures on the previous two datasets, misses most gesture windows on CABB-S, resulting in a higher precision but a low recall.

The two multi-phase models are clearly better than both groups. MPSL \cite{ghaleb2024co} reaches .617 window F1 with the highest accuracy (.928)\footnote{The F1 score reported in \cite{ghaleb2024co} is .672, which is the \textit{average} test F1 of the models from all epochs, according to the released implementation. Our re-run did not replicate this value. For a fair comparison with the other models, we report the performance of the last epoch.}, and our model (CTR-GCN + BiLSTM) reaches the highest window F1, .666 $\pm$ .03, with an accuracy of .918. In particular, our model recovers more gold gesture windows than MPSL (78\% vs.\ 56\% recall).

%% file: Sections/06-analysis.tex
\section{Analysis}
Across all models and datasets, the best gesture F1 does not reach 0.7. Contrary to the optimistic scores reported in prior work (Table~\ref{tab:comparison}), co-speech gesture detection is therefore far from solved. In this section, we examine typical errors.

\begin{figure}[htpb]
    \centering
\includegraphics[width=0.25\linewidth]{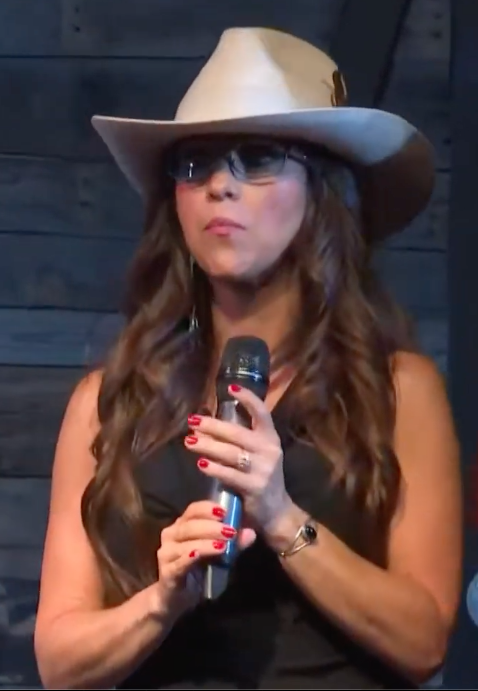}
\includegraphics[width=0.35\linewidth]{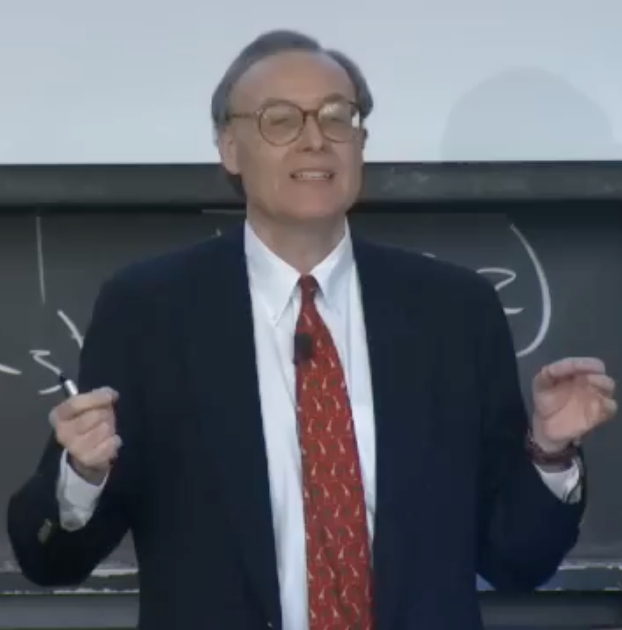}
    \caption{``Non-gestures" detected by all methods as gestures}
\label{fig:gesture_FP_example}
\end{figure}

\begin{figure}[htpb]
    \centering
    \includegraphics[width=0.25\linewidth]{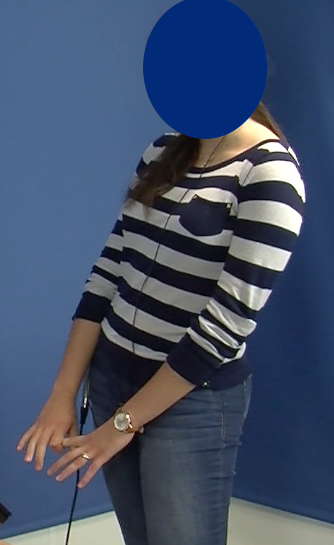}
    \includegraphics[width=0.25\linewidth]{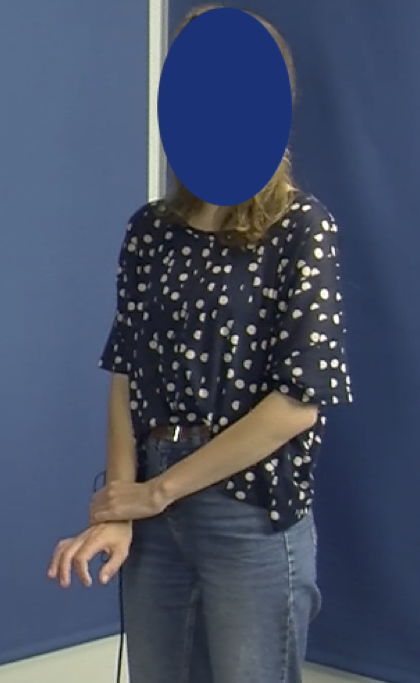}
    \caption{Subtle hand gestures missed by all methods except our proposed method}
    \label{fig:gesture_FN_example}
\end{figure}

Figure~\ref{fig:gesture_FP_example} shows two frames from GESRes that all compared models detect as gestures, although they are not annotated as such (false positives; Table~\ref{tab:gesres_metrics}). On the left, the speaker
is not gesturing, but because she is holding a microphone, her pose resembles a gesture in the pose-tracking data. On the right, the speaker holds his hands in mid-air at the end position of his previous gesture. Depending on the annotation guidelines, and on how consistently annotators
follow them, such a hold may or may not be annotated as part of the preceding gesture; here it was not, yet all models detect it as a gesture.
Conversely, Figure~\ref{fig:gesture_FN_example} shows two CABB-S participants gesturing subtly with their hands while their arms are held low in front of the body, a typical rest position. Most models miss these spans;
only our CTR-GCN-based model detects them. All in all, these examples illustrate the ambiguity and challenge of the task.

\begin{figure}[htpb]
    \centering
    \includegraphics[width=0.8\linewidth]{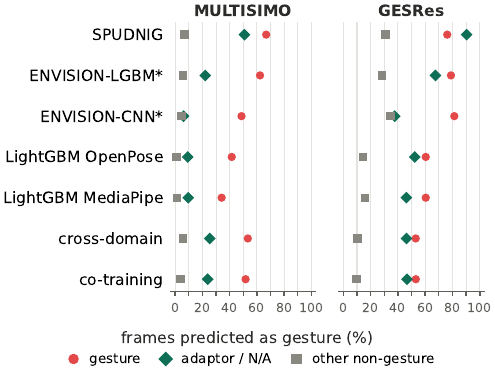}
    \caption{Share of different gold frames predicted as gesture by each method. Non-gesture movement is annotated as \texttt{N/A} in MULTISIMO (
    20,358 frames; 2.6\% of the corpus) and as \texttt{adaptor} in GESRes (
    2,700 frames; 0.7\% of the corpus). $^*$The test data partly overlaps with training data.}
    \label{fig:adaptor}
\end{figure}

Another likely source of error are adaptors, such as adjusting clothing or a microphone. Figure~\ref{fig:adaptor} shows the proportion of gesture,
adaptor and other non-gesture frames that each model labels as gesture. The models follow a similar pattern: they label more adaptor frames as gestures on GESRes than on MULTISIMO, largely in line with the overall rate at which they predict gestures on each dataset. The exception is ENVISION-CNN, which is the only method that labels adaptors as gesture hardly more often than other non-gesture frames 
It is the only method that uses adaptor annotations during training, by learning a separate class for non-gesture movement. This suggests that adaptors are distinguishable from gestures given suitable training labels, and that future work should make use of adaptor annotations where available.

%% file: Sections/07-conclusion.tex
\section{Conclusion}
\label{sec:conclusion}
We introduced a unified framework for evaluating co-speech gesture detection methods fairly: rule-based tools, pre-trained detectors and trained models are run on the same cross-validation folds of three datasets and scored with the same frame-, span- and window-level metrics. Our experiments showed that 1) the success reported for existing methods could not be replicated, and their performance varies strongly with the dataset, specifically with how frequent gestures are; 2) models trained on the target data remain the most reliable, while cross-domain training and co-training help only as far as the training data matches the test data; and 3) our proposed CTR-GCN + BiLSTM model outperforms the previous best method, MPSL, on CABB-S, 
particularly by recovering more gestures.

Identifying gesturing spans appears to be a simple binary task, yet even models trained specifically on the target data reach only moderate F1. The earlier optimism may therefore come partly from evaluation practice rather than from the methods themselves, since each was tested on its own data, with its own unit of evaluation and its own gesture ratio. This matters for how the tools are meant to be used. They were not proposed to replace human annotation but to speed it up, by pointing annotators to candidate gestures. Yet where gestures are rare, most of the candidates they return are not gestures (SPUDNIG: a span precision of 28\% on MULTISIMO and a window precision of 14\% on CABB-S), so they do little to filter the material an annotator has to inspect.

Also, motion alone may not be sufficient. Beat-like gestures are defined largely by their alignment with speech, and whether a movement has a communicative role may only be clear from the speech context. Adding speech, however, is a design choice that depends on how the annotations are used: if the detected gestures are later used to analyze gesture-speech synchronization, a detector that favours speech-aligned gestures could produce biased data.

The main application of gesture detection is annotation for the study of human behaviour, but it could also serve the evaluation of gesture generation. Co-speech gesture generation systems are still mostly evaluated through subjective human comparisons, and the inconsistency and limitations of these
evaluation methods are increasingly being discussed \cite{nagy2026towards}.
A reliable gesture detector and annotator could provide an objective, categorical measure of generated gestures, for example whether a system produces gestures at a plausible rate and at the right moments.

A more interpretable representation of movement may help to understand why detection breaks down across datasets. In future work, we will explore learned discrete pose representations, such as a Vector Quantized Variational Autoencoder (VQ-VAE)~\cite{van2017neural}, which would allow us to inspect which movement patterns a detector treats as gesture and to test whether these representations transfer better across datasets and annotation schemes.